\documentclass[10pt,twocolumn,letterpaper]{article}

\usepackage{cvpr}              
\usepackage{indentfirst}
\usepackage{multirow}
\usepackage{amsmath}
\usepackage{amssymb}
\usepackage{pifont}
\usepackage[dvipsnames]{xcolor}
\usepackage[accsupp]{axessibility}

\newcommand{\xmark}{{\color{red}\ding{55}}}
\newcommand{\cmark}{{\color{OliveGreen}\ding{51}}}
\usepackage[dvipsnames]{xcolor}

\definecolor{cvprblue}{rgb}{0.21,0.49,0.74}
\usepackage[pagebackref,breaklinks,colorlinks,citecolor=cvprblue]{hyperref}

\def\paperID{*****} 
\def\confName{CVPR}
\def\confYear{2024}

\title{GenFlow: Generalizable Recurrent Flow for 6D Pose Refinement of Novel Objects}

\author{Sungphill Moon* \qquad Hyeontae Son* \qquad Dongcheol Hur \qquad Sangwook Kim\\
NAVER LABS\\
{\tt\small \{sungphill.moon, son.ht, dongcheol.hur, o.s.w.a.l.k\}@naverlabs.com}
}

\begin{document}
\maketitle
\let\thefootnote\relax\footnotetext{*These authors contributed equally.}
\begin{abstract}
Despite the progress of learning-based methods for 6D object pose estimation, the trade-off between accuracy and scalability for novel objects still exists.
Specifically, previous methods for novel objects do not make good use of the target object's 3D shape information since they focus on generalization by processing the shape indirectly, making them less effective.
We present GenFlow, an approach that enables both accuracy and generalization to novel objects with the guidance of the target object's shape.
Our method predicts optical flow between the rendered image and the observed image and refines the 6D pose iteratively.
It boosts the performance by a constraint of the 3D shape and the generalizable geometric knowledge learned from an end-to-end differentiable system.
We further improve our model by designing a cascade network architecture to exploit the multi-scale correlations and coarse-to-fine refinement.
GenFlow ranked first on the unseen object pose estimation benchmarks in both the RGB and RGB-D cases.
It also achieves performance competitive with existing state-of-the-art methods for the seen object pose estimation without any fine-tuning.
\end{abstract}
\section{Introduction}
\label{sec:introduction}

Estimating accurate 6D object pose from an RGB or RGB-D image is crucial for robotic tasks and augmented reality applications.
In the standard setup of 6D object pose estimation, 3D models of target objects are given and available in training and testing.
Most state-of-the-art algorithms are learning-based and require training the model per object or dataset containing few objects.
However, these methods are difficult to handle novel objects since they memorize the 3D shapes of the targets and suffer from the training cost to support them.

Recent works \cite{DBLP:conf/icra/OkornGHH21, labbe2022megapose, chen2023zeropose, Shugurov_2022_CVPR} have proposed generalizable models that can estimate the 6D pose of novel objects, \textit{i.e.}, objects unseen during training.
Specifically, OSOP \cite{Shugurov_2022_CVPR} and MegaPose \cite{labbe2022megapose} show remarkable performance with the generalizable pose refinement.
Their refinement methods have the scalability to novel objects since the networks learn to compare the rendered image and the observed image without memorizing 3D shapes.
However, their refiners avail themselves of 3D shape information indirectly.
OSOP refiner is trained by the surrogate 2D-2D matching loss, which makes it suboptimal for 6D pose estimation.
MegaPose refiner regresses the relative 6D pose between the rendered image and the input, but the regression makes no use of the projective geometry.

\begin{figure}[t]
	\centering
	\includegraphics[width=\linewidth]{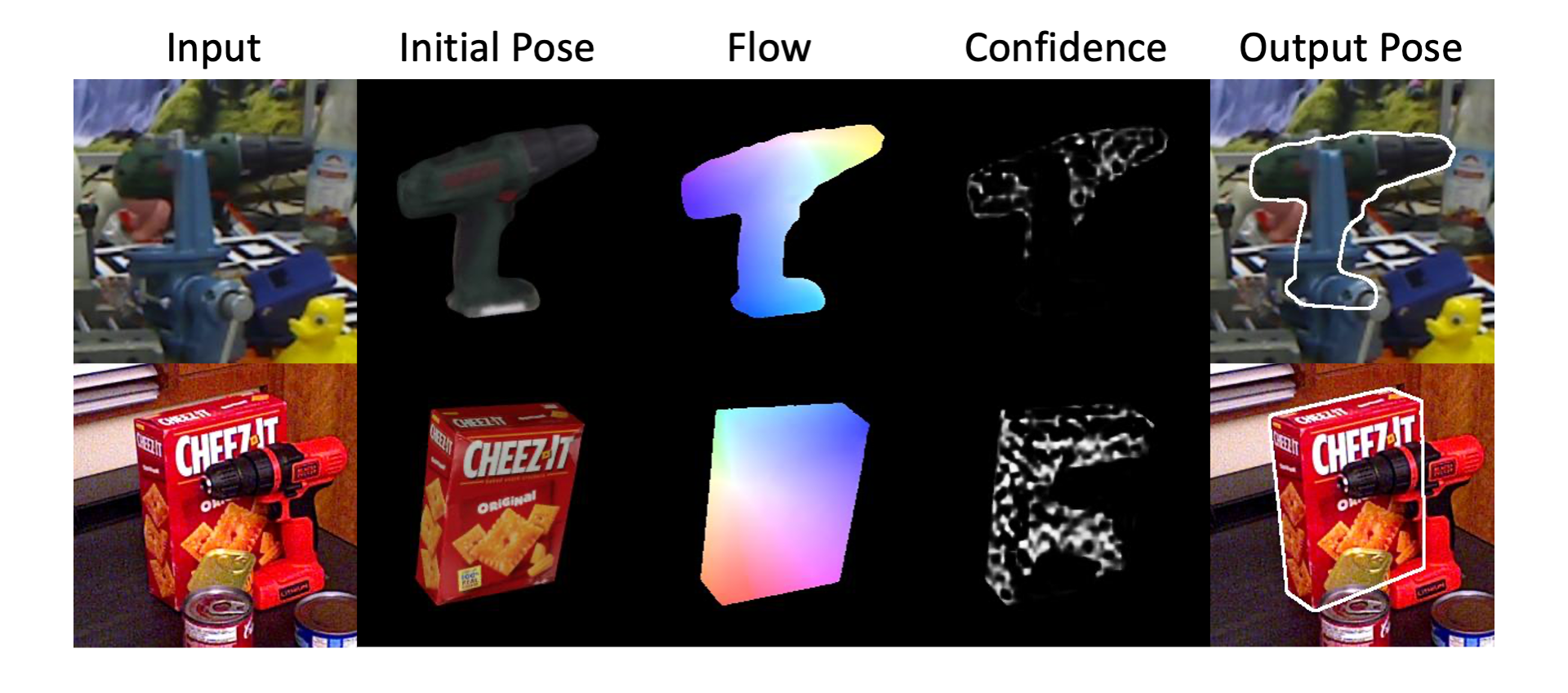}

	\caption{Examples of 6D pose estimation of novel objects. Our method estimates correspondences between the input image and rendered image, and 6D object pose.}
	\label{fig:thumbnail}
\end{figure}

Our research question is \textit{Is there a better design for generalizable pose refinement that takes advantage of shape information?}.
Recently, 2D optical flow-based methods \cite{hu2022pfa, Hai_2023_CVPR} show noticeable performance on 6D pose estimation.
These methods estimate dense 2D-2D correspondences between the rendered image and the observed image.
The estimated correspondences are lifted to 2D-3D correspondences on the 3D shape of the target object, then the 6D pose is recovered using a PnP solver \cite{hu2022pfa} or pose regressor networks \cite{Hai_2023_CVPR}.
The optical flow-based pose refinement provides a clue for the generalization due to its robustness to the domain shift \cite{hu2022pfa} and shows the potential to utilize 3D information in the model \cite{Hai_2023_CVPR}.

We propose GenFlow, the optical flow-based iterative refinement for the generalizable object pose estimation guided by the 3D shape information of target objects.
Given an image and initial 6D pose of an object in the image, GenFlow estimates 2D dense correspondences between the rendered image and the input image and refined 6D pose through a PnP layer.
This series of processes is performed iteratively, so it can improve accuracy and robustness to outliers \cite{Lipson_2022_CVPR}.

We adopt the shape-constraint recurrent flow framework inspired by \cite{Hai_2023_CVPR} for our model to use 3D information.
This framework utilizes the shape information implicitly to enjoy both generalization and performance.
Its end-to-end differentiable system learns to estimate the optical flow, confidence, and object pose.
These estimations are updated iteratively complying with the 3D shape.
We further introduce a cascade network design to make use of the multi-scale correlation volumes.
It takes advantage of the coarse-to-fine refinement and helps the model estimate a more accurate pose.

We validate our method on various datasets provided in the BOP challenge \cite{DBLP:conf/eccv/HodanSDLBMRM20}.
Our model is designed to estimate 6D pose for the RGB input but is also applicable to the RGB-D case with the RANSAC-Kabsch algorithm \cite{kabsch1976solution, DBLP:journals/cacm/FischlerB81}.
For both RGB and RGB-D cases, our method achieves the overall best performance for the unseen object pose estimation task using the default detection \cite{nguyen_2023_cnos} provided in the BOP challenge 2023.
It is also notable that our method achieves competitive performance with the state-of-the-art methods for seen objects when used with better detection results.

\section{Related Work}
\label{sec:related_work}

\noindent \textbf{Iterative Refinement} DeepIM \cite{li2018deepim} pioneered the iterative render-and-compare strategy for 6D pose estimation.
It predicts the relative pose that better matches a rendered image against the observed image.
By iterative refinement, the rendered and observed images become more similar.
CosyPose \cite{labbe2020} improved the idea by introducing better parameterization for rotation \cite{Zhou_2019_CVPR}, network architecture \cite{DBLP:conf/icml/TanL19}, loss design \cite{Wang_2019_CVPR, Simonelli_2019_ICCV} and data augmentation.
These iterative methods show accurate results and the possibility of being applied to the unseen objects \cite{li2018deepim}.
In contrast to the previous works, CIR \cite{Lipson_2022_CVPR} introduces a novel differentiable PnP layer to utilize the projective geometry with deep learning.
Unlike CIR, which assumes the RGB-D input, ours focuses on the RGB input.
For the RGB input case, CIR obtains the rendered depth as the substitution of input depth for every flow update, so the inference suffers from time inefficiency.
On the other hand, our refinement method requires fewer renderings.

\noindent \textbf{Optical Flow Estimation} The classical approaches \cite{DBLP:journals/ai/HornS81, DBLP:conf/iccv/BlackA93, DBLP:conf/cvpr/ChenK16} dealt with optical flow estimation as a hand-crafted optimization problem in pursuit of the balance of brightness consistency and motion plausibility.
Deep learning methods \cite{Dosovitskiy_2015_ICCV, Ilg_2017_CVPR, Hui_2018_CVPR, Sun_2018_CVPR, DBLP:conf/nips/YangR19} emerged as promising alternatives of the classical methods, achieving considerable performance with faster inference times.
Recently, RAFT \cite{DBLP:conf/eccv/TeedD20} introduced a recurrent deep network architecture which has shown significant improvement over the previous methods regarding accuracy, efficiency, and generalization.
PFA \cite{hu2022pfa} applied the optical flow estimation to 6D pose refinement for data-limited cases.
Since learning to estimate dense optical flow helps the networks to focus on the lower-level information, it could be robust to domain shift.
Following PFA, SCFlow \cite{Hai_2023_CVPR} further improved the performance by introducing an end-to-end trainable system using a pose regressor and shape-constraint lookup.
Our method capitalizes on a differentiable PnP layer to the framework in order to make use of projective geometry.

\noindent \textbf{Unseen Object Pose Estimation} Category-level 6D pose estimation \cite{Wang_2019_CVPR, manhardt2020cps, chen2020category, DBLP:journals/ral/LiSBLYI22} tackled the case of unseen objects of which category is given in advance.
Since these methods depend on the prior of the 3D shapes belonging to specific category, they struggle with handling the objects of unknown categories.
Recently, several works proposed class-agnostic generalizable pose estimators that can handle unseen objects.
A few works attempted the CAD model-free pose estimation \cite{Park_2020_CVPR, Sun_2022_CVPR, He_2022_CVPR, liu2022gen6d, he2022oneposeplusplus} to cover everyday objects, but the results were only evaluated on datasets without occlusion and cluttered scenes.
\cite{DBLP:conf/bmvc/XiaoQLAM19, Sundermeyer_2020_CVPR, Xiao2020PoseContrast, Nguyen_2022_CVPR} focused on estimating the orientation of novel objects by comparing the deep image features of the rendered image and the observed image.
ZePHyR \cite{DBLP:conf/icra/OkornGHH21} proposed a generalizable scoring function combined with hand-crafted features \cite{DBLP:conf/iccv/Lowe99, drost2010model}. However, it uses the features selectively according to the richness of texture in the datasets.
ZeroPose \cite{chen2023zeropose} estimates 6D pose by hierarchical feature matching which requires the depth input by its nature and pose refinement for better performance.
Like ours, OSOP \cite{Shugurov_2022_CVPR} and MegaPose \cite{labbe2022megapose} proposed CAD-based pose estimation methods that can process RGB and RGB-D inputs.
Specifically, MegaPose achieved state-of-the-art performance by its iterative pose refiner.
The pose refiner leverages the multi-view rendered images containing surface normals for utilizing rich information about the novel object's shape, anchor point, and coordinate system.
Our method refines the pose with the guidance of the target object's shape without rendering multi-view images per refinement step.

\section{Method}
\label{sec:method}

This section describes our methodology for 6D pose estimation of novel objects.
Given an RGB image and 3D models of target objects, which are unseen during training, our goal is to estimate the 6D poses of the objects with respect to the camera.
Our pose estimation pipeline consists of 3 stages: object detection, coarse pose estimation, and pose refinement.
Note that the detection of novel objects is out of the scope of this paper.
We first provide an overview of the whole process and then focus on the pose refinement method, which is the main contribution of this work.

\subsection{Preliminaries}
Given a calibrated RGB image $\mathcal{I}$ and a set of 3D models of the target objects $\{ \mathcal{M}_{i} \}$, we estimate the 6D poses of the visible objects in the image.
A 6D pose is defined by a matrix ${\mathbf{P}}$ representing the rigid transformation from the object space to the camera space.
Note that we assume the input image is calibrated, \textit{i.e.}, the intrinsic camera matrix ${\mathbf{K}}$ is known.

\noindent \textbf{2D Object Detection} Given an RGB image, the object detection method produces a list of object detections consisting of 2D bounding boxes and associated labels.
For the unseen object detection, we apply the CNOS \cite{nguyen_2023_cnos} detection leveraging the foundation models \cite{kirillov2023segany, oquab2023dinov2}.
To show the extensive results of our work, we also apply the YOLOX detector \cite{DBLP:journals/corr/abs-2107-08430} from GDRNPP \cite{liu2022gdrnpp_bop} which is used for seen object detection.
For each detection, the image $\mathcal{I}$ is cropped and further resized to the specific resolution in the following steps.
The intrinsic parameters are adjusted according to the resized resolution.
For convenience, the following explanation focuses on the 6D pose estimation of a single detected object of which the 3D model is $\mathcal{M}$.

\begin{figure}[t]
	\centering
	\includegraphics[width=\linewidth]{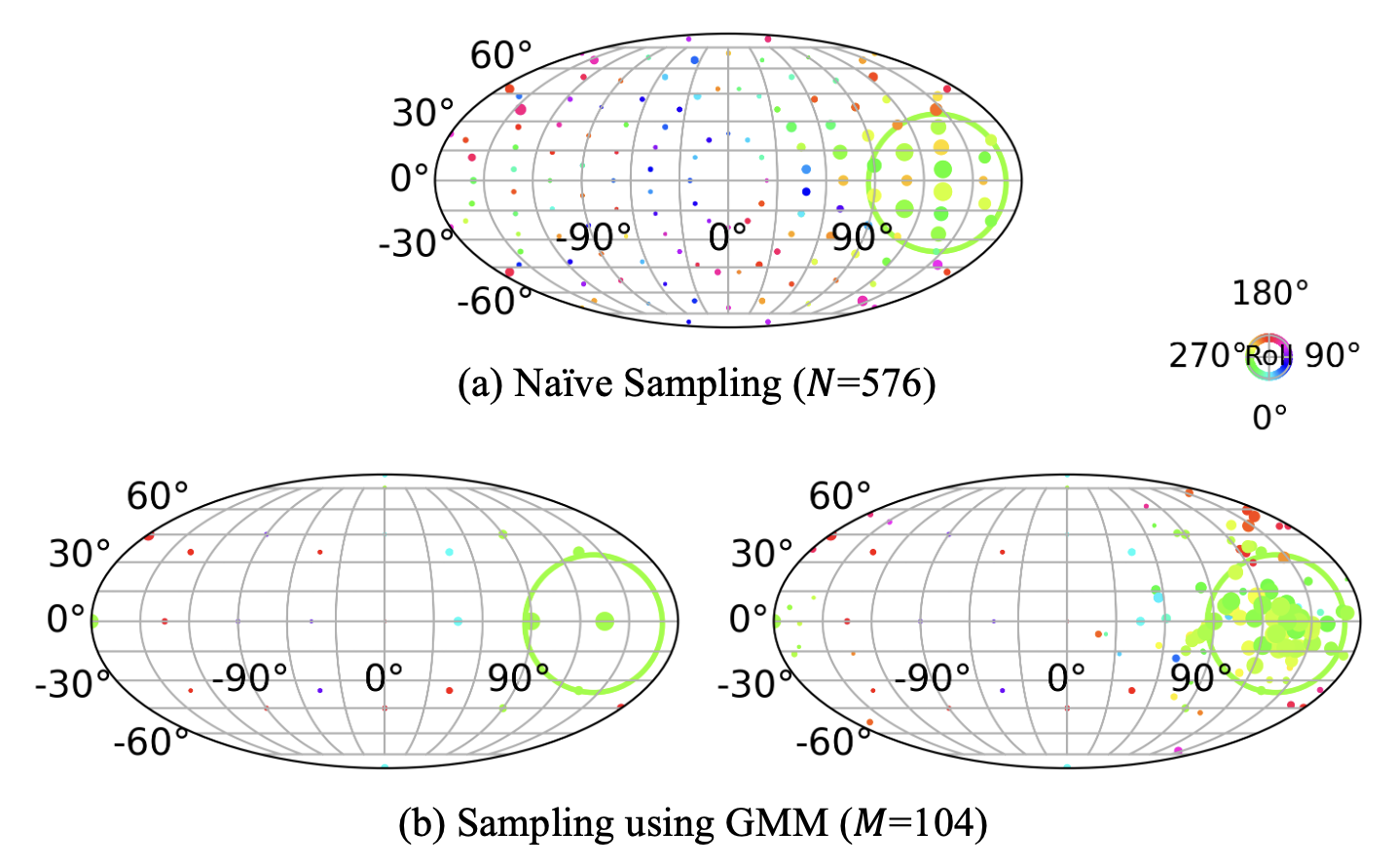}

	\caption{Visualization of the scores predicted by the coarse model. Each dot denotes the sampled rotation, and its size represents the relative score of the coarse model for the corresponding rotation. The green circles are boundaries of the area where the refiner is interested. Figures show that the high-scoring poses are clustered in the vicinity of the green circles.}
	\label{fig:coarse_score}
\end{figure}

\noindent \textbf{Coarse Pose Estimation} Given object detection, the coarse pose estimator predicts rough initial poses.
We used the classification-based method proposed in \cite{labbe2022megapose} for better generalization.
Given an input image and an arbitrary pose $\mathbf{P}$ for the 3D model $\mathcal{M}$, the coarse model classifies whether or not the pose $\mathbf{P}$ could be improved to the pose in proximity to the ground-truth by the refiner.
It uses a pre-defined set of 3D rotations of which the cardinality is $N$ to generate the various arbitrary pose hypotheses.
For each rotation, a rough 3D translation is estimated for the points of the object's 3D model to be approximately inside the 2D bounding box.
We extract scores of the pose hypotheses from the model and select the top-$n$ scoring poses.
Such a classification-based estimation can handle novel objects, but it suffers from the cost of rendering a large number of images for the pre-defined rotations.
Inspired by the property of high-scoring rotations to be sparse and clustered as shown in Fig \ref{fig:coarse_score}, we introduce sampling using a Gaussian mixture model (GMM) \cite{DBLP:books/lib/Bishop07} for efficiency.
First, we choose the top $k$-scoring samples ${ \{ R_{i} = (\psi_{i}, \theta_{i}, \phi_{i} )\}}_{i=1}^{k}$ from $M$ pre-defined rotations (Fig \ref{fig:coarse_score}b left).
Then we create a density distribution $p(X)$ for the Euler angle $X$ using a GMM which satisfies that
\begin{align}
	p(X) = \sum_{i=1}^{k} {\frac{1}{k} \mathcal{N}(X|\mu_{i}, \Sigma)}, \mu_{i}=R_{i}
\end{align}
Finally, we sample additional $M$ rotations from the density distribution $p(X)$ (Fig \ref{fig:coarse_score}b right) and choose the best-scoring $n$ hypotheses among the $2M$ samples.
This method renders $2M < N$ images while achieving better performance.
We used $M=104$ and $k=16$ for our experiments.

\noindent \textbf{Pose Refinement} Given a coarse pose estimate, we refine the pose for the rendered image to match against the input image crop.
Starting from the coarse estimate $\mathbf{P}_{0}$ as an initial pose, we iteratively refine the pose using our refiner.
We denote the $k^{th}$ 6D pose estimate of the refiner as $\mathbf{P}_{k}$.
For the ${k^{th}}$ iteration, we obtain an image ${\mathcal{I}_{r,k}}$ by rendering the 3D model $\mathcal{M}$ for the pose $\mathbf{P}_{k-1}$.
Our refiner then estimates the 2D optical flow ${\mathbf{F}_{k}}$ between the image ${\mathcal{I}_{r,k}}$ and the input image crop and the confidence weights $\mathbf{W}_{k}$ which are used to solve the pose ${\mathbf{P}_{k}}$.
A detailed description is in the following section.

\begin{figure*}[t]
	\centering
	\includegraphics[width=\linewidth]{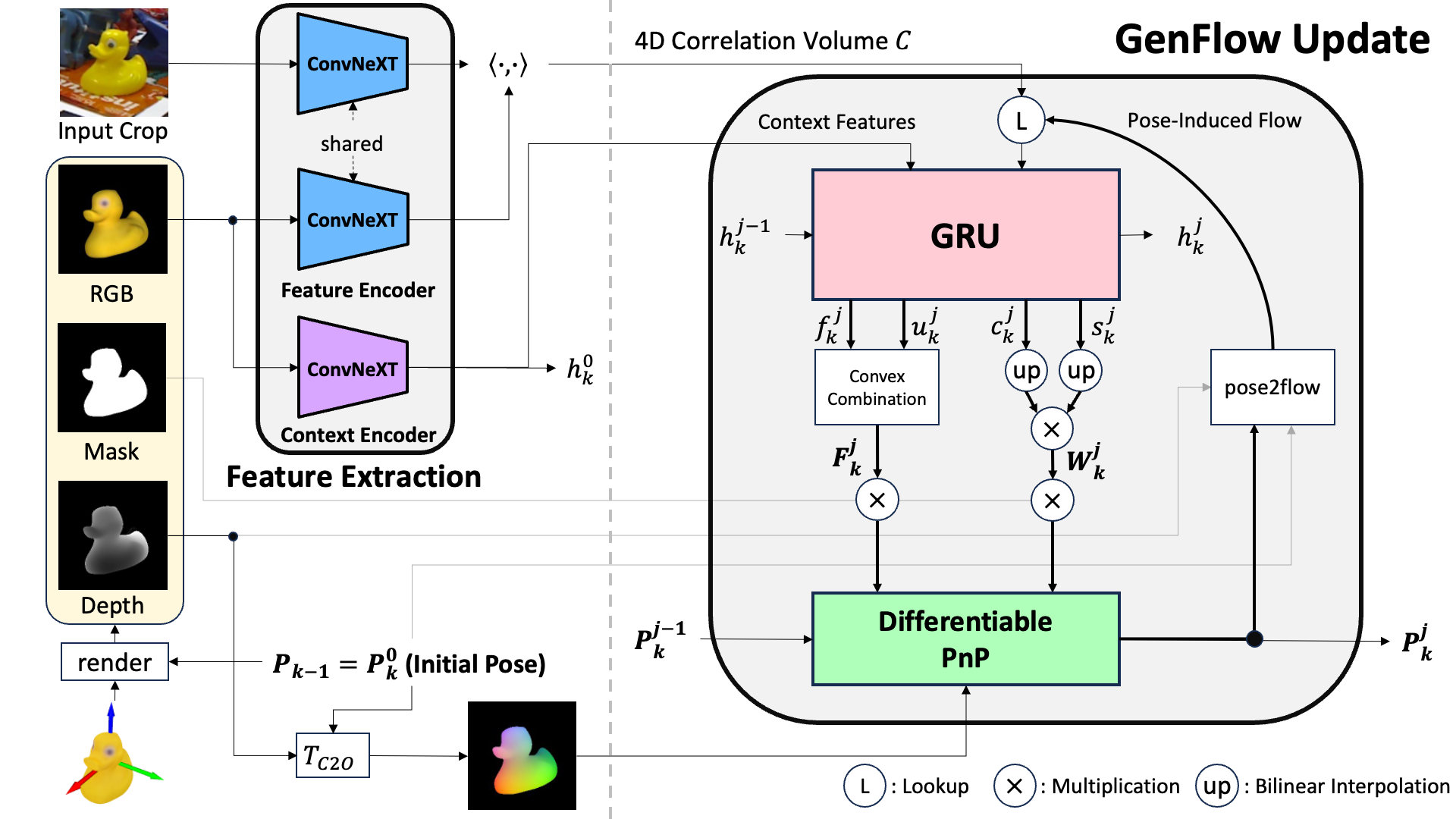}

	\caption{Overview of GenFlow refinement. We visualize the process of feature extraction and GenFlow update in the $k^{th}$ refinement. ${T}_{C2O}$ refers to a transformation that maps the coordinates from camera space to object space.}
	\label{fig:refiner_structure}
\end{figure*}
\subsection{GenFlow Refinement}
\subsubsection{Overview}
We describe the specific operation of GenFlow refiner to estimate the refined pose $\mathbf{P}_{k}$ from the previous result $\mathbf{P}_{k-1}$ but hereafter, we omit the $k$ notation if possible for convenience.
First, we obtain a synthetic image by rendering the 3D model $\mathcal{M}$ for the pose $\mathbf{P}$ and adjusted intrinsic parameters.
Specifically, the results of the rendering consists of the RGB image, depth image and binary rendered mask, denoted by ${\mathcal{I}_{r} \in R^{H \times W \times 3} }$, ${\mathcal{D}_{r} \in R^{H \times W}}$, and ${\mathcal{M}_{r} \in \{0,1\}^{H \times W}}$ respectively.
Then, we extract features from both images and refine the pose with the GenFlow module.

\noindent \textbf{Feature Extraction} We extract the image features from the input image crop and the rendered image ${\mathcal{I}_{r}}$ with the same feature encoder and the context features from the ${\mathcal{I}_{r}}$.
The feature maps from the $i^{th}$ layer of the feature encoder are used to construct the correlation volumes ${\mathbf{C}_{i} \in R^{H_{i} \times W_{i} \times H_{i} \times W_{i}}}$ by the dot product.

\noindent \textbf{GenFlow Module} For each correlation volume and context feature map, we construct a module named ``GenFlow module".
A GenFlow module consists of a convolution-based Gated Recurrent Unit (GRU) \cite{DBLP:conf/ssst/ChoMBB14}, a differentiable PnP solver, and correlation lookup based on the pose-induced flow as shown in the Fig \ref{fig:refiner_structure}.
GenFlow module iteratively updates the optical flow, confidence weights, and 6D pose.
For the $j^{th}$ update, the output of the module is the refined 6D pose $\mathbf{P}^{j}$.
Note that we use subscripts for the outer updates and superscripts for the inner updates, \textit{i.e.}, GenFlow updates.
The detailed illustration of updating the 6D pose with a GenFlow module for ${i}^{th}$ correlation volume is as follows.

\subsubsection{GenFlow Update}
\begin{figure}[t]
	\centering
	\includegraphics[width=\linewidth]{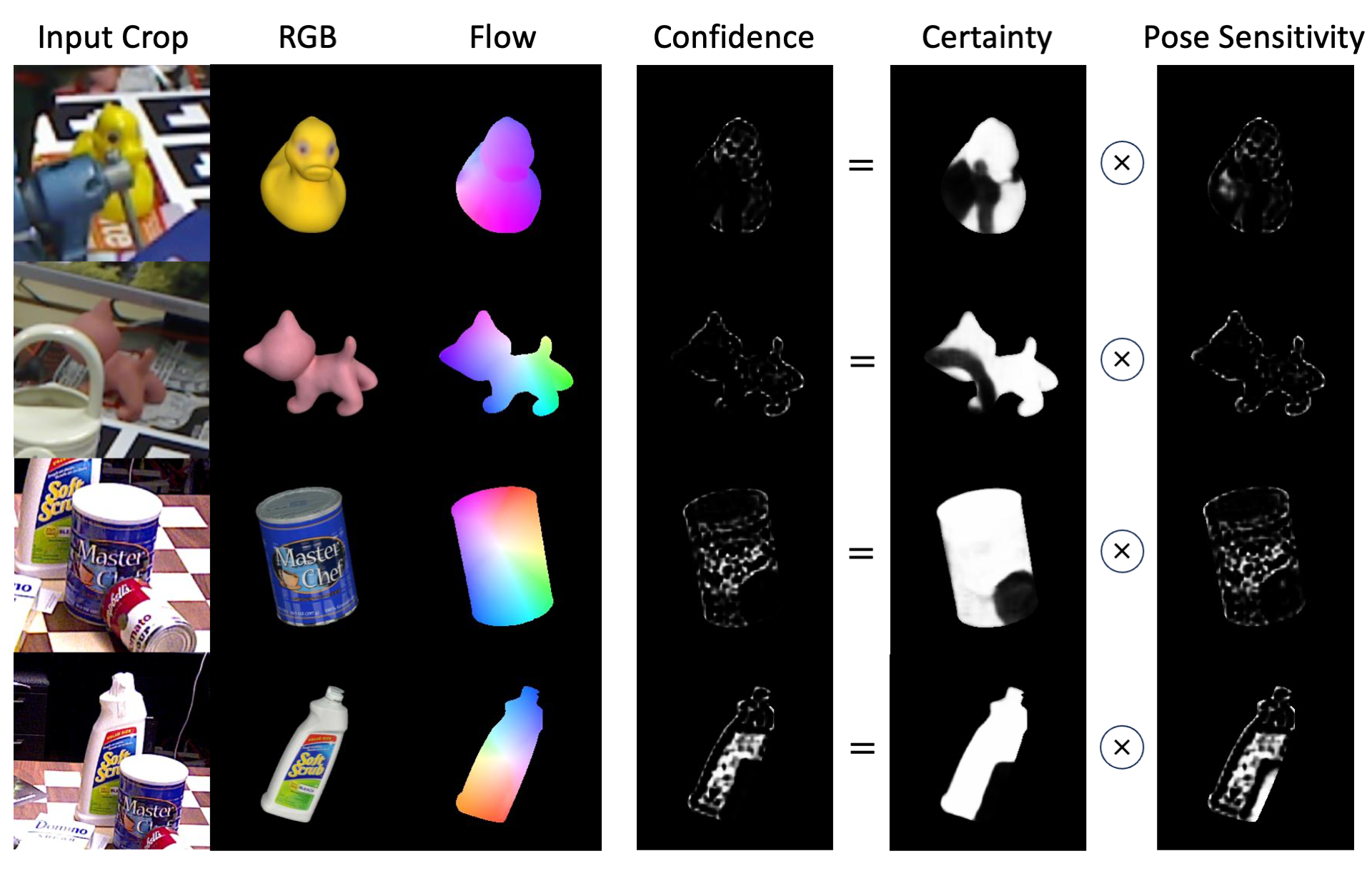}

	\caption{Visualization of the outputs of GRU. Our method factorizes the confidence weights into two terms: certainty and pose sensitivity. Certainty estimation helps to the robustness to occlusion. Pose sensitivity highlights the rich texture regions and extremities of the object.}
	\label{fig:outputs_visualization}
\end{figure}

\noindent \textbf{GRU Update} For the $j^{th}$ update, the GRU of the GenFlow module consumes the previous hidden state $h^{j-1}$, correlation features ${r^{j-1}}$ and context features.
The output of the GRU $\{\mathbf{h}^{j}, \mathbf{f}^{j}, \mathbf{u}^{j}, \mathbf{c}^{j}, \mathbf{s}^{j}\}$ includes the updated hidden state $h^{j}$, optical flow ${\mathbf{f}^{j}}$, flow upsampling mask ${\mathbf{u}^{j}}$, certainty ${\mathbf{c}^{j}}$ and pose sensitivity ${\mathbf{s}^{j}}$.
The flow ${\mathbf{f}^{j} \in R^{H_{i} \times W_{i} \times 2}}$ is upsampled to ${\mathbf{F}^{j}}$ to match the resolution of previous ${(i-1)}^{th}$ feature map using the mask $\mathbf{u}^{j}$.
We apply the convex combination method of 3x3 neighbors for the optical flow proposed in \cite{DBLP:conf/eccv/TeedD20}.
The certainty $\mathbf{c}^{j}$ and pose sensitivity ${\mathbf{s}^{j}}$ are upsampled by bilinear interpolation, and then we multiply them to obtain the confidence weights ${\mathbf{W}^{j}}$.
The head networks for certainty and pose sensitivity are supervised by different losses, respectively.
This explicit factorization of the confidence weights helps to be robust to occlusion with the help of certainty estimation as shown in Fig \ref{fig:outputs_visualization}.
The upsampled flow $\mathbf{F}^{j}$ and confidence weights $\mathbf{W}^{j}$ are used to establish weighted 2D-3D correspondences.

\noindent \textbf{Pose Update} As we know the depth of synthetic image $\mathcal{D}_{r}$ for the pose $\mathbf{P}_{k}^{0}=[\mathbf{R}_{k}^{0}|\mathbf{t}_{k}^{0}]$ and the intrinsic camera matrix $\mathbf{K}$, we can compute the object-space 3D coordinate of each pixel by
\begin{align}
	\left(
	\begin{matrix}
		x \\
		y \\
		z
	\end{matrix}
	\right) =
	\mathbf{R}_{k}^{\text{T}} (\mathbf{K}^{-1} \mathcal{D}_{r}(u, v)
	\left(
	\begin{matrix}
		u \\
		v \\
		1
	\end{matrix}
	\right) - \mathbf{t}_{k})
	\label{eq:lifting}
\end{align}
where $(u, v)$ is the 2D coordinate of a pixel and the $\mathcal{D}_{r}(u, v)$ is depth of the pixel.
Through this lifting process, we obtain the 2D-3D correspondences for the flow $\mathbf{F}^{j}$.
We use $\mathbf{W}^{j}$ for the confidence of correspondences and binary rendered mask $\mathcal{M}_{r}$ to consider the object region only.

To obtain the 6D pose $\mathbf{P}^{j}$, we utilize the iterative PnP solver based on the Levenberg-Marquardt (LM) algorithm.
It optimizes the 6D pose by minimizing the sum of squared weighted reprojection error:
\begin{align}
	\begin{split}
		\operatorname*{argmin}_{R, t} \frac{1}{2} \sum_{u} \sum_{v}
		\lVert
		\mathbf{W}^{j}(u,v) \times
		(\pi(R
		\left(
		\begin{matrix}
			x \\
			y \\
			z
		\end{matrix}
		\right) + t) \\
		- 
		(\left(
		\begin{matrix}
			u \\
			v \\
		\end{matrix}
		\right) + \mathbf{F}^{j}(u, v))
		)
		\rVert^{2},
	\end{split}
\end{align}
where $\pi$ is the projection function with the intrinsic camera parameters and $(x,y,z)^{\text{T}}$ is the 3D coordinate of the pixel $(u,v)$ computed by the equation \ref{eq:lifting}.
Due to its differentiable nature, the PnP layer enables end-to-end training.
However, the backpropagation by unrolling the iterative optimization causes numerical instability, making the training process hard.
For stable end-to-end training, we introduce the regularization method \cite{Chen_2022_CVPR}, which backpropagates the gradient through the last step of a differentiable optimization.
Specifically, we first obtain a detached pose $\mathbf{P}^{*}$ through 3 LM steps and update the pose via another iteration of the Gauss-Newton algorithm to obtain the pose $\mathbf{P}^{j}$.
Note that the gradient is not backpropagated through $\mathbf{P}^{*}$.

\noindent \textbf{Correlation Lookup} We compute the pose-induced flow, which is the 2D optical flow between the $\mathcal{I}_{r}$ and the image with respect to the pose $\mathbf{P}^{j}$.
The pose-induced flow is calculated as the displacement of each pixel in the $\mathcal{I}_{r}$ by lifting and reprojecting the pixel using the depth $\mathcal{D}_{r}$.
To embed the 3D shape of the target object in the lookup operation \cite{Lipson_2022_CVPR, Hai_2023_CVPR}, we use the pose-induced flow rather than the estimated flow $\mathbf{f}^{j}$ for indexing the correlation volume $\mathbf{C}$.
This way of imposing a shape constraint on the model induces the networks to predict dense matching to comply with the target's 3D shape information.
From the lookup operation, we obtain the correlation feature $r^{j}$, which is the input of GRU for the next GenFlow update.

\subsubsection{Cascade Architecture}
\begin{figure}[t]
	\centering
	\includegraphics[width=\linewidth]{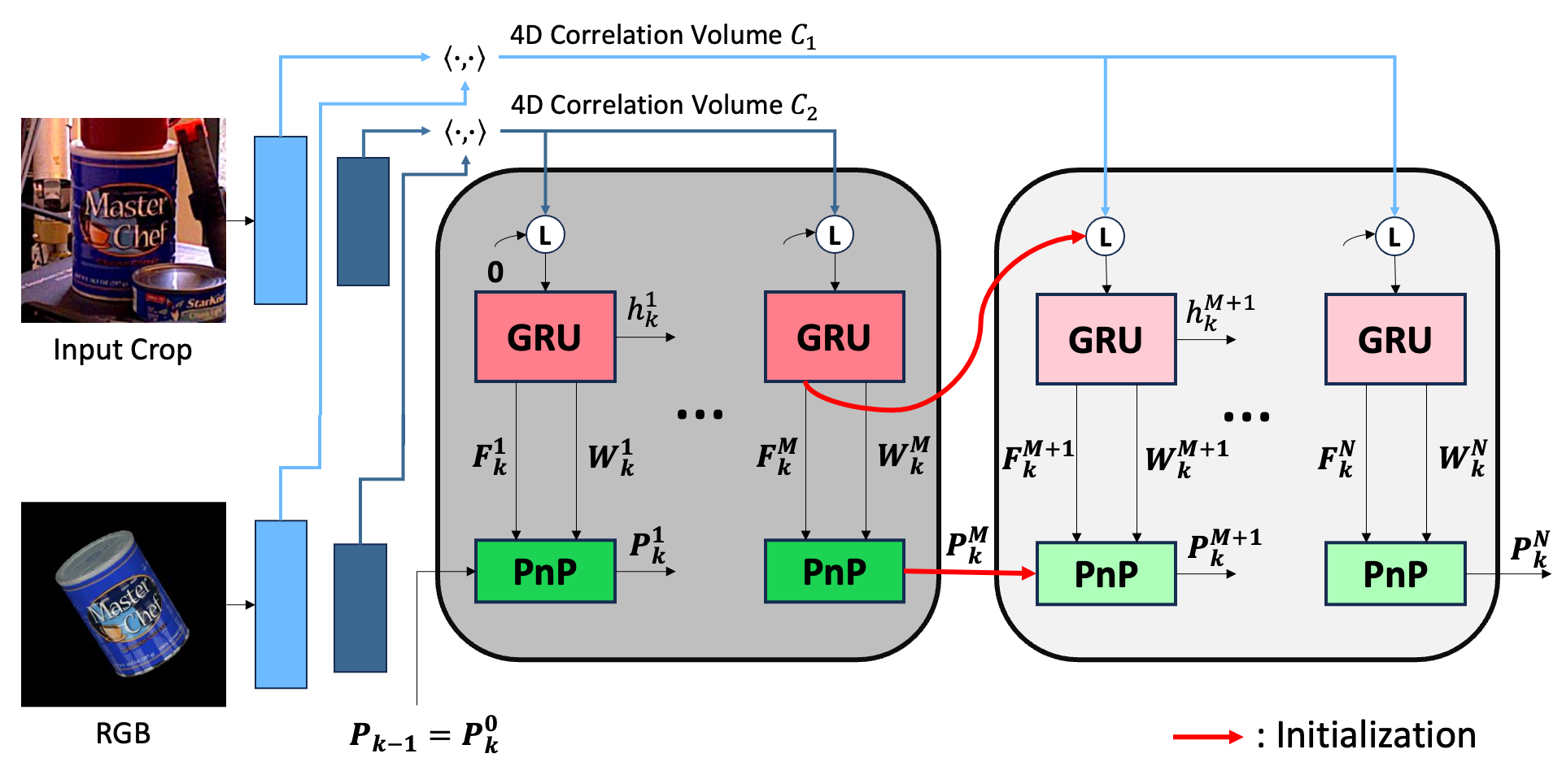}

	\caption{Cascade design of GenFlow modules. Multiple GenFlow modules are attached to each level on the feature pyramid. The last updated flow $\mathbf{F}_{k}^{M}$ and 6D pose $\mathbf{P}_{k}^{M}$ from the higher-level GenFlow module are used to initialize the flow and 6D pose of the lower-level module. With the cascade architecture, the 6D pose is recovered in a coarse to fine manner.}
	\label{fig:cascade_architecture}
\end{figure}

Inspired by the previous methods that exploit the multi-scale features \cite{Lin_2017_CVPR, Chen_2018_CVPR}, we design a cascade architecture using multiple GenFlow modules.
Given correlation volumes ${\{ \mathbf{C}_{i} \}}$, we assign a GenFlow module to each volume.
Then, we use the 2D optical flow and refined 6D pose of the last inner updates to initialize those of the next module as shown in Fig \ref{fig:cascade_architecture}.
This cascade architecture takes advantage of progressive refinement, which makes use of the high-level semantics to the low-level fine details.
We found that it improves the accuracy of 6D pose rather than using a single GenFlow module.

\subsection{Training}
\noindent \textbf{Dataset} Both the coarse and refiner models are trained using the large dataset provided by MegaPose \cite{labbe2022megapose}.
It is a large synthetic dataset comprising 2 million RGB-D images generated by BlenderProc \cite{Denninger2023}, which contains diverse 3D models with ground-truth 6D pose annotations.
The 3D shapes are collected from ShapeNet \cite{DBLP:journals/corr/ChangFGHHLSSSSX15} and Google Scanned Object (GSO) \cite{DBLP:conf/icra/DownsFKKHRMV22} datasets,
and we trained our model on both of the datasets.

\noindent \textbf{Refiner Model} Our refiner model is trained similarly to the previous iterative refinement methods \cite{labbe2020, labbe2022megapose}.
Given an image with the ground-truth 6D pose $\mathbf{P}_{gt}$ for the 3D model $\mathcal{M}$, we perturb the pose to generate an initial pose $\mathbf{P}_{init}$ by applying random noises to the translation and rotation of the pose respectively.
Translation noise is sampled from a normal distribution with a standard deviation of (0.01, 0.01, 0.05) centimeters, and rotation noise is sampled from another normal distribution of Euler angles with a standard deviation of 15 degrees in each axis.
The refiner model learns to predict the pose $\mathbf{P}_{gt}$ from the perturbed initial pose $\mathbf{P}_{init}$.

\noindent \textbf{Coarse Model} Given an input RGB image $\mathcal{I}$ and a pose $\mathbf{P}$, a coarse model is trained to classify whether the refiner could estimate the 6D pose in proximity to the ground truth pose from the initial pose $\mathbf{P}$.
We randomly generate the positive and negative pose samples and train the network to classify them using the binary cross entropy loss.
For the positive samples, we use the same method used to generate the perturbed initial pose in the refiner model.
Negative samples are generated using the method proposed in \cite{labbe2022megapose}.

\subsection{Implementation Details}
\noindent \textbf {Coarse Estimation} We use 224$\times$224 resolution for the resized input crop and the synthetic image $\mathcal{I}_{r}$.
The coarse model is implemented using the ConvNeXt backbone \cite{Liu_2022_CVPR} and fully connected head.
We used the ConvNext model pre-trained on ImageNet \cite{DBLP:conf/cvpr/DengDSLL009} classification to initialize the networks' weights.
For inference, we select top-$n$ hypotheses using our GMM-based sampling strategy with $M=104$.

\noindent \textbf {GenFlow Refinement} First, we obtain a synthetic image $\mathcal{I}_{r}$ of which resolution is 256$\times$256 and an image crop resized to the same resolution as the synthetic image with adjusted intrinsic parameters.
Like the coarse model, we use the pre-trained ConvNeXt for the feature and context encoders.
The feature maps from the first two layers of the feature encoder are used to construct correlation volumes ${\mathbf{C}_{1} \in R^{\frac{H}{4} \times \frac{W}{4}  \times \frac{H}{4} \times \frac{W}{4}}}$ and ${\mathbf{C}_{2} \in R^{\frac{H}{8} \times \frac{W}{8}  \times \frac{H}{8} \times \frac{W}{8}}}$.
Our GRU is implemented with a ConvGRU unit \cite{DBLP:conf/eccv/TeedD20} and following four CNN-based heads for the 2D optical flow $\mathbf{f}$,  flow upsampling mask $\mathbf{u}$, certainty $\mathbf{c}$ and pose sensitivity $\mathbf{s}$.
Two GenFlow modules are attached to the correlation volumes for the cascade architecture.
They are serially connected so that the estimations from higher-level features are utilized as initial values for the next module.

The optical flow and its upsampling mask produce the upsampled flow, which is supervised by the L1 endpoint error loss with the ground-truth optical flow.
We also utilize the disentangled point matching loss for the pose loss, supervising the rotation, 2D object center, and depth of the object individually.
Specifically, given an estimated pose $\mathbf{P}=[\mathbf{R} | [\mathbf{t}_{x}, \mathbf{t}_{y}, \mathbf{t}_z]^{\textnormal{T}}]$ and ground truth pose $\mathbf{P}_{gt}=[\mathbf{R}^{\star} | [\mathbf{t}_{x}^{\star}, \mathbf{t}_{y}^{\star}, \mathbf{t}_z^{\star}]^{\textnormal{T}}]$, the pose loss is computed as follows:
\begin{align}
	\begin{split}
		\mathcal{L}_{pose} = \mathcal{D}([\mathbf{R}|[\mathbf{t}_{x}^{\star}, \mathbf{t}_{y}^{\star}, \mathbf{t}_{z}^{\star}]^{\text{T}}], \mathbf{P}_{gt})  \\
		+ \mathcal{D}([\mathbf{R}^{\star}|[\mathbf{t}_{x}, \mathbf{t}_{y}, \mathbf{t}_{z}^{\star}]^{\text{T}}], \mathbf{P}_{gt}) \\
		+ \mathcal{D}([\mathbf{R}^{\star}|[\mathbf{t}_{x}^{\star}, \mathbf{t}_{y}^{\star}, \mathbf{t}_{z}]^{\text{T}}], \mathbf{P}_{gt})
	\end{split}
\end{align}
where the distance $\mathcal{D}$ between two 6D poses $\mathbf{P}_{1}$ and $\mathbf{P}_{2}$ using the 3D points $\mathcal{X}_{\mathcal{M}}$ on the 3D model $\mathcal{M}$ is defined as
\begin{align}
	\mathcal{D}(\mathbf{P}_{1}, \mathbf{P}_{2}) = \frac{1}{\lvert\mathcal{X}_{\mathcal{M}}|} \sum_{x \in \mathcal{X}_{\mathcal{M}}} |\mathbf{P}_{1}x - \mathbf{P}_{2}x\rvert
\end{align}

To supervise the certainty $\mathbf{c}$, we introduce the certainty estimation by classifying the depth consistency similarly to \cite{Edstedt_2023_CVPR}.
Given 6D poses $\mathbf{P}_{init}$ and $\mathbf{P}_{gt}$ for the 3D model $\mathcal{M}$ and intrinsic camera parameters $\mathbf{K}$, we compute the projected depth $\mathcal{D}_{r \rightarrow t}$ by warping the rendered image to the target input image using the depth $\mathcal{D}_{r}$.
We also obtain the depth map of warped pixels $\mathcal{D}_{t}$ given the synthetic depth of the target image.
Then we train the networks by minimizing the binary cross entropy loss $\mathcal{L}_{cert}$ between $\mathbf{c}$ and $\mathbf{c}_{r \rightarrow t}$ which is a binary mask $\mathbf{c}_{r \rightarrow t} = | \mathcal{D}_{r \rightarrow t} - \mathcal{D}_{t} | < d_{th}$ where $d_{th}$ is the distance threshold.
It is worth noting that we place a stop-gradient on the certainty mask $\mathbf{c}^{j}$ when computing the pose loss $\mathcal{L}_{pose}^{j}$ for disentangling the certainty $\mathbf{c}$ and pose sensitivity $\mathbf{s}$.

Given the RGB-D images with annotated ground truth 6D poses, the overall loss is defined as
\begin{align}
	\mathcal{L} = \sum_{j=1}^{N} \gamma^{j-N}(\mathcal{L}_{flow}^{j}  + \alpha \mathcal{L}_{cert}^{j} + \beta \mathcal{L}_{pose}^{j})
\end{align}
following the strategy of exponentially increasing weights \cite{DBLP:conf/eccv/TeedD20} where we use the weight $\gamma=0.8$ and the iteration of GenFlow updates $N=8$ in this work.
In specific, the GenFlow module assigned to $\mathbf{C}_{2}$ outputs $\{\mathbf{P}^{j}\}_{j=1}^{N/2}$ and the another module outputs $\{\mathbf{P}^{j}\}_{j=N/2}^{N}$ from $\mathbf{C}_{1}$.

\makeatletter
\newcommand\newtag[2]{#1\def\@currentlabel{#1}\label{#2}}
\makeatother
\begin{table*}[]
    \setlength{\tabcolsep}{0.7pt}
    \newcommand{\colorizeTableRow}{\color{magenta}}
    {\footnotesize
    \begin{tabular}{rc|cc|cc|cc|cccccccc}
        \hline
                                                            &                                                       & \multicolumn{2}{c|}{\multirow{2}{*}{2D Localization}}                                                                       & \multicolumn{2}{c|}{\multirow{2}{*}{Pose Initialization}}                                         & \multicolumn{2}{c|}{\multirow{2}{*}{Pose Refinement}}                                                                                      & \multicolumn{8}{c}{\multirow{2}{*}{Average Recall ($\uparrow$)}}                                                                                                                                        \\
                                                            & \multicolumn{1}{l|}{}                                 & \multicolumn{2}{c|}{}                                                                                                       & \multicolumn{2}{c|}{}                                                                             & \multicolumn{2}{c|}{}                                                                                                                      & \multicolumn{8}{c}{}                                                                                                                                                                                    \\ \hline
                                                            & {\scriptsize{\begin{tabular}[c]{@{}c@{}}RGB-D\\ Input\end{tabular}}} & Method                                                            & {\scriptsize{\begin{tabular}[c]{@{}c@{}}Novel\\ Objects\end{tabular}}} & Method                                  & {\scriptsize{\begin{tabular}[c]{@{}c@{}}Novel\\ Objects\end{tabular}}} & Method                                                                           & {\scriptsize{\begin{tabular}[c]{@{}c@{}}Novel\\ Objects\end{tabular}}} & {\scriptsize{LM-O}} & {\scriptsize{T-LESS}} & {\scriptsize{TUD-L}} & {\scriptsize{IC-BIN}} & {\scriptsize{ITODD}} & {\scriptsize{HB}} & \multicolumn{1}{c|}{{\scriptsize{YCB-V}}} & {\scriptsize{MEAN}} \\ \hline
        {\scriptsize{\colorizeTableRow{\newtag{1}{1.1}}}}   & \xmark                                                & Mask-RCNN \cite{DBLP:conf/iccv/HeGDG17, labbe2020}                & \xmark                                                  & CosyPose \cite{labbe2020}               & \xmark                                                  & CosyPose \cite{labbe2020}                                                        & \xmark                                                  & 63.3                & 64.0                  & 68.5                 & 58.3                  & 21.6                 & 65.6              & \multicolumn{1}{c|}{57.4}                 & 57.0                \\
        {\scriptsize{\colorizeTableRow{\newtag{2}{1.2}}}}   & \xmark                                                & Mask-RCNN \cite{DBLP:conf/iccv/HeGDG17, labbe2020}                & \xmark                                                  & SurfEmb \cite{Haugaard_2022_CVPR}       & \xmark                                                  & BFGS                                                                             & \xmark                                                  & 66.3                & 73.5                  & 71.5                 & 58.8                  & 41.3                 & 79.1              & \multicolumn{1}{c|}{64.7}                 & 65.0                \\
        {\scriptsize{\colorizeTableRow{\newtag{3}{1.3}}}}   & \xmark                                                & YOLOX \cite{DBLP:journals/corr/abs-2107-08430, liu2022gdrnpp_bop} & \xmark                                                  & ZebraPose \cite{Su_2022_CVPR}           & \xmark                                                  & -                                                                                & \xmark                                                  & \textbf{72.9}       & 81.1                  & 75.6                 & 59.2                  & \textbf{50.4}        & \textbf{92.1}     & \multicolumn{1}{c|}{\textbf{72.9}}        & \textbf{72.0}       \\
        {\scriptsize{\colorizeTableRow{\newtag{4}{1.4}}}}   & \xmark                                                & YOLOX \cite{DBLP:journals/corr/abs-2107-08430, liu2022gdrnpp_bop} & \xmark                                                  & MegaPose \cite{labbe2022megapose}       & \cmark                                                  & MegaPose+MH \cite{labbe2022megapose}                                             & \cmark                                                  & 64.8                & 78.1                  & 74.1                 & 56.9                  & 42.2                 & 86.3              & \multicolumn{1}{c|}{70.2}                 & 67.5                \\
        {\scriptsize{\colorizeTableRow{\newtag{5}{1.5}}}}   & \xmark                                                & YOLOX \cite{DBLP:journals/corr/abs-2107-08430, liu2022gdrnpp_bop} & \xmark                                                  & Ours                                    & \cmark                                                  & Ours+MH                                                                          & \cmark                                                  & 68.3                & \textbf{82.8}         & \textbf{77.8}        & \textbf{59.6}         & 50.1                 & 89.7              & \multicolumn{1}{c|}{70.8}                 & 71.3                \\ \hline
        {\scriptsize{\colorizeTableRow{\newtag{6}{1.6}}}}   & \cmark                                                & YOLOX \cite{DBLP:journals/corr/abs-2107-08430, liu2022gdrnpp_bop} & \xmark                                                  & WDR-Pose \cite{DBLP:conf/cvpr/HuSJFS21} & \xmark                                                  & PFA \cite{hu2022pfa}+Kabsch                                                      & \xmark                                                  & \textbf{79.2}       & \textbf{84.9}         & \textbf{96.3}        & \textbf{70.6}         & 52.6                 & 86.7              & \multicolumn{1}{c|}{\textbf{89.9}}        & \textbf{80.0}       \\
        {\scriptsize{\colorizeTableRow{\newtag{7}{1.7}}}}   & \cmark                                                & YOLOX \cite{DBLP:journals/corr/abs-2107-08430, liu2022gdrnpp_bop} & \xmark                                                  & MegaPose \cite{labbe2022megapose}       & \cmark                                                  & MegaPose+MH \cite{labbe2022megapose}+Teaserpp \cite{DBLP:journals/trob/YangSC21} & \cmark                                                  & 70.4                & 71.8                  & 91.6                 & 59.2                  & 55.3                 & 87.2              & \multicolumn{1}{c|}{85.5}                 & 74.4                \\
        {\scriptsize{\colorizeTableRow{\newtag{8}{1.8}}}}   & \cmark                                                & YOLOX \cite{DBLP:journals/corr/abs-2107-08430, liu2022gdrnpp_bop} & \xmark                                                  & Ours                                    & \cmark                                                  & Ours+MH+Kabsch                                                                   & \cmark                                                  & 74.2                & 78.3                  & 92.8                 & 64.9                  & \textbf{65.2}        & \textbf{92.0}     & \multicolumn{1}{c|}{88.3}                 & 79.4                \\ \hline
        {\scriptsize{\colorizeTableRow{\newtag{9}{1.9}}}}   & \xmark                                                & OSOP \cite{Shugurov_2022_CVPR}                                    & \cmark                                                  & OSOP \cite{Shugurov_2022_CVPR}          & \cmark                                                  & OSOP+PnP+MH \cite{Shugurov_2022_CVPR}                                            & \cmark                                                  & 31.2                & -                     & -                    & -                     & -                    & 49.2              & \multicolumn{1}{c|}{33.2}                 & -                   \\
        {\scriptsize{\colorizeTableRow{\newtag{10}{1.10}}}} & \xmark                                                & CNOS-det. \cite{nguyen_2023_cnos}                                 & \cmark                                                  & MegaPose \cite{labbe2022megapose}       & \cmark                                                  & MegaPose+MH \cite{labbe2022megapose}                                             & \cmark                                                  & 56.0                & 50.8                  & 68.7                 & 41.9                  & 34.6                 & 70.6              & \multicolumn{1}{c|}{62.0}                 & 54.9                \\
        {\scriptsize{\colorizeTableRow{\newtag{11}{1.11}}}} & \xmark                                                & CNOS-det. \cite{nguyen_2023_cnos}                                 & \cmark                                                  & Ours                                    & \cmark                                                  & Ours+MH                                                                          & \cmark                                                  & \textbf{57.5}       & \textbf{53.0}         & \textbf{69.1}        & \textbf{45.6}         & \textbf{40.8}        & \textbf{74.5}     & \multicolumn{1}{c|}{\textbf{63.9}}        & \textbf{57.8}       \\ \hline
        {\scriptsize{\colorizeTableRow{\newtag{12}{1.12}}}} & \cmark                                                & OSOP \cite{Shugurov_2022_CVPR}                                    & \cmark                                                  & OSOP \cite{Shugurov_2022_CVPR}          & \cmark                                                  & OSOP+Kabsch+MH \cite{Shugurov_2022_CVPR}+ICP                                     & \cmark                                                  & 48.2                & -                     & -                    & -                     & -                    & 60.5              & \multicolumn{1}{c|}{57.2}                 & -                   \\
        {\scriptsize{\colorizeTableRow{\newtag{13}{1.13}}}} & \cmark                                                & CNOS-seg. \cite{nguyen_2023_cnos}                                 & \cmark                                                  & ZeroPose \cite{chen2023zeropose}        & \cmark                                                  & MegaPose+MH \cite{labbe2022megapose}                                             & \cmark                                                  & 53.8                & 40.0                  & 83.5                 & 39.2                  & 52.1                 & 65.3              & \multicolumn{1}{c|}{65.3}                 & 57.0                \\
        {\scriptsize{\colorizeTableRow{\newtag{14}{1.14}}}} & \cmark                                                & CNOS-det. \cite{nguyen_2023_cnos}                                 & \cmark                                                  & MegaPose \cite{labbe2022megapose}       & \cmark                                                  & MegaPose+MH \cite{labbe2022megapose}+Teaserpp \cite{DBLP:journals/trob/YangSC21} & \cmark                                                  & 62.6                & 48.7                  & 85.1                 & 46.7                  & 46.8                 & 73.0              & \multicolumn{1}{c|}{76.4}                 & 62.8                \\
        {\scriptsize{\colorizeTableRow{\newtag{15}{1.15}}}} & \cmark                                                & CNOS-det. \cite{nguyen_2023_cnos}                                 & \cmark                                                  & Ours                                    & \cmark                                                  & Ours+Kabsch+MH                                                                   & \cmark                                                  & \textbf{62.9}       & \textbf{51.7}         & \textbf{85.8}        & \textbf{53.3}         & \textbf{55.9}        & \textbf{78.2}     & \multicolumn{1}{c|}{\textbf{82.5}}        & \textbf{67.2}       \\ \hline
    \end{tabular}
    }
    \caption{6D pose estimation results on the BOP challenge datasets. We report the Average Recall (AR) scores across the datasets for various methods. The higher score the better. The best results among the comparable methods are in bold. We denote the multi-hypotheses strategy as MH for simplicity.}
    \label{table:main_result}
\end{table*}

\noindent \textbf{Multi-Hypotheses Strategy} We select $n$ best-scoring hypotheses in the coarse estimation.
The selected hypotheses are refined, and we choose the best estimate by evaluating the refined hypotheses using the coarse model again.
This strategy requires more inference time for evaluating $n$ iterations of the rendering and refinement process, but it can boost the accuracy by avoiding the risk of the ``Winner-Takes-All" strategy.

\noindent \textbf{RGB-D Input} If the observed image consists of depth information, we apply the depth refinement after the GenFlow loop for every refinement step.
First, we obtain the optical flow $\mathbf{F}$ and upsampled certainty from the results of the last GenFlow update.
3D-3D correspondences are established by the optical flow and the input depth.
We filter out the correspondences of certainty lower than a specific threshold to remove the outliers.
Then we apply the RANSAC-Kabsch \cite{kabsch1976solution, DBLP:journals/cacm/FischlerB81} for the depth refinement.
Using certainty for filtering out the outliers is more effective than using confidence weights since the high-confidence pixels are sparse and the input depth is prone to be noisy.

\section{Experiments}
\label{sec:experiments}

\subsection{Datasets and Evaluation Metrics}
\noindent \textbf{Datasets} We evaluate our method on seven datasets of the BOP challenge \cite{DBLP:conf/eccv/HodanSDLBMRM20}: LM-O \cite{DBLP:conf/accv/HinterstoisserLIHBKN12}, T-LESS \cite{DBLP:conf/wacv/HodanHOMLZ17}, TUD-L \cite{hodan2018bop}, IC-BIN \cite{DBLP:conf/cvpr/DoumanoglouKMK16}, ITODD \cite{DBLP:conf/iccvw/DrostUBHS17}, HomebrewdDB \cite{DBLP:conf/iccvw/KaskmanZSI19} and YCB-V \cite{xiang2018posecnn}.
These datasets contain cluttered real-world scenes captured from different lighting and camera sensors.
Each scene consists of multiple objects including occlusions between them.
The objects vary in terms of texture richness, symmetry, and usage.

\noindent \textbf{Evaluation Metrics} Following the evaluation methodology of BOP challenge \cite{DBLP:conf/eccv/HodanSDLBMRM20}, we report the average recall (AR) considering three pose-error functions: Visible Surface Discrepancy (VSD), Maximum Symmetry-Aware Surface Distance (MSSD), and Maximum Symmetry-Aware Projection Distance (MSPD).
VSD measures the misalignment of the visible part by computing the ratio of pixels in which the difference between the estimated distance to the camera center and ground truth is under a specific tolerance.
In consideration of the global symmetry information, MSSD computes the maximum distance between the estimated camera-space coordinates and corresponding ground truth. In contrast, MSPD computes the maximum distance between the estimated image-space coordinates and the corresponding ground truth.
Concerning each pose-error function, the AR is obtained by averaging recall calculated for multiple settings of the correctness thresholds (and misalignment tolerances for VSD).

\subsection{BOP Benchmark Results}
Table \ref{table:main_result} shows the results of our method on the BOP datasets.
For a fair comparison, all reported pose initialization and pose refinement methods are trained with synthetic data containing images rendered with BlenderProc \cite{Denninger2023}.
We reported the AR scores of MegaPose and ours by setting the number of hypotheses $n$ as 10 for the multi-hypotheses strategy and the number of outer updates as 5.
For ours, the number of GenFlow updates per single outer update is set to 8.

\noindent \textbf{Unseen Objects} For both RGB input (row \ref{1.9}-\ref{1.11}) and RGB-D input (row \ref{1.12}-\ref{1.15}), the results show that our approach outperforms all other methods for novel objects.

\noindent \textbf{Seen Objects} From row \ref{1.1} to \ref{1.8}, the pose estimation results on the 2D detection trained for target datasets are reported.
Our method still outperforms MegaPose for the same detection results.
It is noticeable that our method provides competitive results to the state-of-the-art methods, though it is not the best, even without fine-tuning on the target datasets.

\begin{table}[]
    \setlength{\tabcolsep}{2.2pt}
    {\footnotesize
    \begin{tabular}{cccccccc}
        \hline
        \multirow{2}{*}{Method} & \multirow{2}{*}{\# of renderings ($ \downarrow $)} & \multicolumn{6}{c}{Average Recall ($ \uparrow $)}                                                                                                            \\ \cline{3-8}
                                &                                                   & {\scriptsize{LM-O}} & {\scriptsize{T-LESS}} & {\scriptsize{TUD-L}} & {\scriptsize{IC-BIN}} & \multicolumn{1}{c|}{{\scriptsize{YCB-V}}} & {\scriptsize{MEAN}} \\ \hline
        Na\"ive                 & 576                                               & 22.5                & 25.6                  & 30.6                 & 17.5                  & \multicolumn{1}{c|}{16.4}                 & 22.5                \\ \hline
        \multirow{2}{*}{Ours}   & 144                                               & 28.8                & 35.5                  & 33.8                 & 29.7                  & \multicolumn{1}{c|}{28.9}                 & 31.3                \\
                                & 208                                               & \textbf{32.9}       & \textbf{40.4}         & \textbf{40.0}        & \textbf{31.1}         & \multicolumn{1}{c|}{\textbf{34.7}}        & \textbf{35.8}       \\ \hline
    \end{tabular}
    }
    \caption{Ablation study of the hypotheses generation strategy for the coarse estimation. The AR scores are reported for the best pose hypothesis from the candidates from different sampling methods.}
    \label{table:gmm_result}
\end{table}
\begin{table*}[]
    \setlength{\tabcolsep}{3.5pt}
    \newcommand{\colorizeTableRow}{\color{magenta}}
    \begin{tabular}{r|l|l|cccccc}
        \hline
                                                            & \multirow{2}{*}{Method}                                & \multirow{2}{*}{Training}                                        & \multicolumn{6}{c}{Average Recall ($ \uparrow $)}                                                                                                            \\ \cline{4-9}
                                                            &                                                        &                                                                  & {\scriptsize{LM-O}} & {\scriptsize{T-LESS}} & {\scriptsize{TUD-L}} & {\scriptsize{IC-BIN}} & \multicolumn{1}{c|}{{\scriptsize{YCB-V}}} & {\scriptsize{MEAN}} \\ \hline
        {\footnotesize{\colorizeTableRow{\newtag{1}{3.1}}}} & No shape-constraint + RANSAC-PnP \cite{Lepetit:160138} & $\mathcal{L}_{flow}$                                             & 61.7                & 77.6                  & 72.4                 & 54.2                  & \multicolumn{1}{c|}{65.7}                 & 66.3                \\ \hline
        {\footnotesize{\colorizeTableRow{\newtag{2}{3.2}}}} & Shape-constraint                                       & $\mathcal{L}_{flow}$, $\mathcal{L}_{pose}$                       & 64.9                & 81.8                  & 75.1                 & 57.0                  & \multicolumn{1}{c|}{70.1}                 & 69.8                \\
        {\footnotesize{\colorizeTableRow{\newtag{3}{3.3}}}} & Shape-constraint + Cascade                             & $\mathcal{L}_{flow}$, $\mathcal{L}_{pose}$                       & 65.7                & 82.3                  & 75.0                 & 58.9                  & \multicolumn{1}{c|}{69.9}                 & 70.4                \\ \hline
        {\footnotesize{\colorizeTableRow{\newtag{4}{3.4}}}} & Shape-constraint + Confidence factorization            & $\mathcal{L}_{flow}$, $\mathcal{L}_{cert}$, $\mathcal{L}_{pose}$ & 64.5                & 81.8                  & 75.3                 & 57.3                  & \multicolumn{1}{c|}{70.5}                 & 69.9                \\
        {\footnotesize{\colorizeTableRow{\newtag{5}{3.5}}}} & Shape-constraint + Confidence factorization + Cascade  & $\mathcal{L}_{flow}$, $\mathcal{L}_{cert}$, $\mathcal{L}_{pose}$ & 65.9                & 82.0                  & 76.1                 & 59.5                  & \multicolumn{1}{c|}{69.8}                 & \textbf{70.7}       \\ \hline
    \end{tabular}
    \caption{Ablation study of GenFlow design. The best result is in bold. Our method accomplishes the best performance (row \ref{3.5}).}
    \label{table:ablation_genflow}
\end{table*}
\subsection{Ablation Study}
We conduct ablation studies to validate the efficacy of our method.
The ablation results are reported for the RGB inputs we mainly focus on.
All experiments were performed on 5 BOP datasets: LM-O, T-LESS, TUD-L, IC-BIN, and YCB-V.
We used the 2D detection results of YOLOX \cite{DBLP:journals/corr/abs-2107-08430, liu2022gdrnpp_bop} trained on target datasets to focus on the accuracy of 6D pose estimation by preventing the performance bottleneck caused by the detection methods for novel objects.

\subsubsection{Coarse Pose Estimation}

Table \ref{table:gmm_result} shows the effectiveness of the GMM-based sampling method for generating pose hypotheses.
Pre-defined rotations are generated following \cite{DBLP:journals/ijrr/YershovaJLM10} for the Na\"ive sampling and for creating GMM of our method.
We used the $N=576$ rotations for the Na\"ive sampling, and $M=72$ and $M=104$ which conduct $2M$ renderings (144 and 208, respectively) for ours.
The results show that our method can generate a better initial pose despite the fewer renderings and network inferences.

\subsubsection{GenFlow design}

Table \ref{table:ablation_genflow} shows the results on different refiner designs.
All experiments are conducted using the same result of coarse estimation, which uses our GMM-based sampling with $M=104$ without the multi-hypotheses strategy.
We run 5 iterations of GenFlow refinement, each containing the 8 iterations of the inner update.
We evaluated the final 6D pose from the GenFlow refiner.

\noindent \textbf{Shape-Constraint} Learning to estimate 2D correspondences with flow loss only is suboptimal for 6D pose estimation since it makes no use of the target shape (row \ref{3.1}).
Imposing shape-constraint on the model benefits from the guide of 3D information without loss of generalization performance (row \ref{3.2}-\ref{3.5}).

\noindent \textbf{Cascade Architecture} The proposed cascade architecture with multiple GenFlow modules contributes to the performance improvement (row \ref{3.3}, \ref{3.5}).

\noindent \textbf{Confidence Factorization} Factorizing the confidence weights to certainty and pose sensitivity helps to increase the accuracy of the output pose.
Certainty estimation leverages the training depth data and makes our method robust to occlusion (row \ref{3.4}, \ref{3.5}).

\subsubsection{MegaPose Refiner vs Ours}
\begin{figure}[t]
	\centering
	\includegraphics[width=\linewidth]{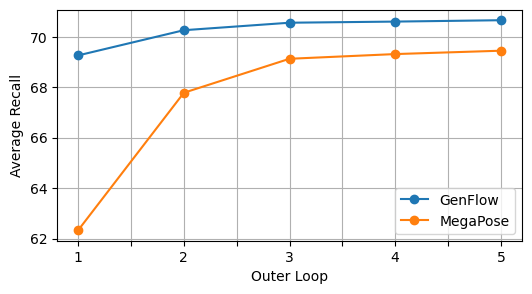}

	\caption{GenFlow vs MegaPose refiner for the RGB inputs.}
	\label{fig:iteration_exp}
\end{figure}

We compare the performance of the MegaPose refiner and ours for the RGB input.
The same pose initializations from our coarse estimation were used for a fair evaluation.
We reported the changes in AR score according to the iteration of the outer updates for both methods in Fig \ref{fig:iteration_exp}.
The reported score is computed as the arithmetic mean of the AR scores for the five datasets.
A single outer update takes 66.5 milliseconds for the MegaPose refiner and 98.2 milliseconds for ours, respectively, on an RTX 3090 GPU and Core i9-10920X CPU.
Although ours is slower than the MegaPose refiner for a single refinement step, it achieves better performance for the same and even fewer iterations.
Note that ours performs effectively by comparing a single-view rendered image for an outer update, while MegaPose refiner uses multiple rendered views as input.

\section{Conclusion}
\label{sec:conclusion}

In this paper, we present GenFlow, an optical flow-based method for 6D pose refinement of novel objects.
First, we generate pose hypotheses using a GMM-based sampling strategy for efficient and effective coarse estimation.
Then, we refine the high-scoring hypotheses with our pose refiner, our main contribution.
GenFlow iteratively refines the flow and confidence with the guidance of 3D shape.
Our method achieves state-of-the-art performance of 6D pose estimation for novel objects concerning RGB and RGB-D inputs.
{
    \small
    \bibliographystyle{ieeenat_fullname}
    \bibliography{main}
}


\end{document}